\documentclass[fleqn,10pt]{wlscirep}
\usepackage[utf8]{inputenc}
\usepackage[T1]{fontenc}
\usepackage{colortbl}
\usepackage{pgfplots}
\usepackage{array}
\usepackage{subcaption}
\usepackage[]{geometry}

\title{A monthly sub-national Harmonized Food Insecurity Dataset for comprehensive analysis and predictive modeling}

\author[1,$\dag$,*]{Melissande Machefer}
\author[1,$\dag$,*]{Michele Ronco}
\author[2]{Anne-Claire Thomas}
\author[1]{Michael Assouline}
\author[3]{Melanie Rabier}
\author[1]{Christina Corbane}
\author[1]{Felix Rembold}
\affil[1]{Joint Research Center, European Commission, Ispra, 21027, Italy}
\affil[2]{Action Against Hunger, Regional Office for West and Central Africa, Dakar, 29621, Senegal}
\affil[3]{IPC, Global Support Unit, Rome, 00153, Italy now at OCHA, Centre for Humanitarian Data, The Hague, 2511 CJ, Netherlands}
\affil[*]{corresponding author(s): Melissande Machefer (melissande.machefer@ec.europa.eu), Michele Ronco (michele.ronco@ec.europa.eu)}

\affil[$\dag$]{these authors contributed equally to this work}

\begin{abstract}
Food security is a complex, multidimensional concept challenging to measure comprehensively. Effective anticipation, monitoring, and mitigation of food crises require timely and comprehensive global data. This paper introduces the Harmonized Food Insecurity Dataset (HFID), an open-source resource consolidating four key data sources: the Integrated Food Security Phase Classification (IPC)/Cadre Harmonisé (CH) phases, the Famine Early Warning Systems Network (FEWS NET) IPC-compatible phases, and the World Food Program's (WFP) Food Consumption Score (FCS) and reduced Coping Strategy Index (rCSI). Updated monthly and using a common reference system for administrative units, the HFID offers extensive spatial and temporal coverage. It serves as a vital tool for food security experts and humanitarian agencies, providing a unified resource for analyzing food security conditions and highlighting global data disparities. The scientific community can also leverage the HFID to develop data-driven predictive models, enhancing the capacity to forecast and prevent future food crises.
\end{abstract}
\begin{document}

\flushbottom
\maketitle

\thispagestyle{empty}

\section*{Background \& Summary}

Achieving the United Nations Sustainable Development Goals 2 -- preventing food shortages and reducing global hunger --  is an imperative endeavor \cite{SDG23}. A critical step toward this ambitious goal is the establishment of continuous and reliable monitoring of food security situation, striving for the broadest geographic coverage and the highest temporal frequency possible. Anticipating and understanding the evolution of food security is essential for effective policy-making. Despite the need for consistently updated data, especially in remote and often most vulnerable areas, this objective remains challenging. Significant efforts have been made by various institutions and networks towards providing the essential data needed to inform strategies that can preempt and alleviate food crises and with tools to visualise \cite{machefer2024reviewML4FS} and monitor the situation. Notable among these are the campaigns and methods supported by the Cadre Harmonisé (CH), the Food and Agriculture Organization (FAO), the Famine Early Warning Systems Network (FEWS NET),  the Integrated Food Security Phase Classification (IPC), the World Bank (WB), and the World Food Programme (WFP). 

However, monitoring food insecurity at a global level faces significant challenges due to the uneven coverage of countries and regions. Moreover, disparate criteria for measuring food insecurity are employed, depending on the goals of the stakeholders. The complexity of food insecurity, influenced by context specific factors, challenges the development of a universal, objective measure, due to its inherent diversity and multi-dimensionality \cite{cafiero2014}. The frequency of data collection and the periods monitored of the various food security indicators often vary, reflecting the differing needs of each initiative at local level. Additionally, geographical inconsistencies arise from the use of various reference systems for administrative boundaries and locations. Consequently, the current landscape of food insecurity assessment is marked by a lack of cohesion, highlighting the need for a more integrated and systematic methodology at a global level \cite{Maxwell2008, Coates2013, Headey2013, Maxwell2014, Allee2021}.

Data scientist practitioners have attempted to leverage publicly available datasets \cite{machefer2024reviewML4FS}, proposing models for nowcasting and forecasting a stand alone target indicator \cite{lentz2019data,andree2020predicting,zhou2022machine,westerveld2021forecasting,deleglise2022food,wang2022transitions,martini2022machine,krishnamurthy2022anticipating,foini2023forecastability,busker2023predicting,herteux2023forecasting}. Using multiple proxies and targets to assess food security would be more effective than a single measure, as it yields a fuller, multidimensional understanding of food insecurity. However, some food insecurity datasets remain unused, due to recent availability, but also the complexity of the definition and of the technical direct access. 

To bridge these gaps, we have developed the Harmonized Food Insecurity Dataset (HFID), a comprehensive compilation of major available sub-national food insecurity data sources with updates on a sub-annual basis. Our dataset spans from $2007$ to the present and encompasses four key variables: the IPC/CH phases \cite{IPC}, the FEWS NET-IPC compatible phases \cite{FEWSNET} and prevalences of population with insufficient food consumption \cite{machefer2024reviewML4FS} based on single indicators of food consumption: the Food Consumption Score (FCS) and reduced Coping Strategy Index (rCSI), both sourced from WFP \cite{WFP} through two different methods. Records within the dataset correspond to the first or second administrative level of the countries, harmonized using the Global Administrative Areas (GADM) database as the standard reference for the nomenclature of administrative units \cite{gadm}. The HFID comprises $311,838$ records, encompassing $5,508$ units and $1,264$ units respectively corresponding to the second and first administrative division across $80$ countries. By integrating these variables within a unified temporal and spatial framework, we enable a thorough and unparalleled comparative analysis of food insecurity trends derived from various datasets.

The HFID not only provides the ability to examine historical data sourced from different providers but also stands as a powerful instrument for ongoing monitoring of food insecurity using both assessments from multi-partner analytical approaches (e.g. IPC/CH) and single outcome indicators (e.g. FCS). It presents a centralized hub for accessing a range of crucial variables, thus streamlining data management and improving the precision and impact of data interpretation. Additionally, it serves as a vital tool for identifying data availability and critically, the areas where data is lacking, encouraging more targeted data collection efforts in underrepresented areas or countries. Finally, the HFID stands as a valuable resource in its own right while also offering significant potential to propel forward the development of advanced modeling techniques within the field of food security. Most studies have targeted the prediction of single indicators. There seems to be an absence of efforts to create forecasts that combine various food insecurity indices, utilizing either multi-output models or the construction of an aggregate target index \cite{Caccavale2020}. It would therefore be intriguing to enrich the HFID's target variables with predictors that encompass the recognized drivers of food crises, in order to associate different food insecurity targets to the varying impacts of these predictors\cite{ipcaccuracystudy}. This could indicate a multitude of underlying mechanisms at play and of food insecurity dimensions. Exploring these applications, we expect a considerable improvement in our understanding of food insecurity and the complex interplay between elements like weather, climate, conflict, and other disruptive events.

\section*{Methods}

Several food insecurity raw data sources have been used to build the HFID. We collected the IPC/CH and FEWS NET-IPC compatible current situation phases through the freely accessible APIs \cite{api_fews, api_ipc}, respectively referred hereafter with phase-IPC/CH and phase-FEWS. WFP  prevalence of population with insufficient food consumption based on single indicators FCS and rCSI, were instead retrieved from two different sources: a publicly available historical dataset\cite{wfp_zenodo}) from the Literature (LIT) covering 2006-2021, called in the following WFP-LIT, and a dedicated data dump with all historical records from Real Time (RT) monitoring also available (only last 500 days) in the HungerMap \cite{api_hungermap}, called in the following WFP-RT. We hereafter refer to the population prevalence of insufficient FCS (poor and borderline) and crisis and above rCSI ($\geq 19$)\footnote{These thresholds used for computing population prevalence of insufficient food consumption for FCS and rCSI indicators are the ones used for defining IPC/CH Phase 3 and higher (see the IPC Acute Food Insecurity Reference Table \cite{ipcmanual3}, page 37).} as FCS-LIT and rCSI-LIT for WFP-LIT, and FCS-RT and rCSI-RT for WFP-RT to indicate from which data sources the indicator is retrieved.

Both FEWS NET and IPC/CH provide a unified classification of Acute Food Insecurity (AFI) of a geographic area in the form of five classes: Phase 1 is None/Minimal Food Insecurity, Phase 2 is Stressed, Phase 3 is Crisis, Phase 4 is Emergency, and Phase 5 is  Catastrophe/Famine. In the IPC/CH API, the "overall\_phase" key corresponds to this IPC/CH phase from 1 to 6, with numbers from 1 to 4  matching their corresponding IPC/ CH Phase, while IPC/CH Phase 5  that can take either the value 5  (Famine with solid evidence) or the value 6 represents Famine (with reasonable evidence)\footnote{See IPC API documentation \url{https://observablehq.com/@ipc/ipc-api-extended-documentation}}. In the IPC/CH, to associate a phase to a geographical area, the analysts must use all relevant and reliable evidence on food security for that area and period of analysis. The evidence used results from a combination of outcomes indicators (such as FCS) and other contributing factors (such as the presence of conflicts). The IPC AFI framework identifies food consumption, livelihood change as first order outcome of food security and nutritional status, and, mortality as second order outcome indicators. The IPC/CH AFI analysis protocols require specific proxy indicators on food consumption and livelihood change to be included. The IPC AFI Reference Table contains all the indicators considered as direct evidence and provides comparable cut-offs associated with the five IPC phases for each indicator\cite{ipcmanual3}. The FEWS NET analyses are IPC-compatible, as they utilize IPC protocols, but are not always based on a multi-partner consensus. This scenario may occur when specific organizations request analyses and classifications of food security conditions without a collaborative agreement among technical experts. For example, this might be due to the pressing need for immediate action in response to an urgent food security situation \footnote{See FEWSNET website \url{https://fews.net/about/integrated-phase-classification}}. 

The FCS and rCSI are food consumption indicators widely collected and used by WFP and partners and which are referred in the IPC Acute Food Insecurity Reference Table to document the food consumption outcome \cite{ipcmanual3}. The FCS intends to measure the quality and quantity of household’s food access, however only its association with energy food consumption has been validated  \cite{hoddinott2002dietary}. The FCS aggregates household-level data on dietary diversity, frequency of food consumption and relative nutritional value of food consumed over the last $7$ days \cite{Leroy2015}. Based on the score obtained, a household's food consumption is further classified into three  categories: poor, borderline, or acceptable. The WFP-LIT and RT-LIT data sources provide the sum of the prevalence of households with poor and borderline FCS which is called insufficient  food consumption. The rCSI measures the frequency and severity of behaviors in which people have engaged in the last $7$ days in response to challenges in accessing food \cite{Maxwell2014}.  It aggregates household data on the use of five pre-weighted coping strategies in the last $7$ days. The WFP-LIT and RT-LIT data sources provide the prevalence of households with an rCSI score above 19 \cite{martini2022machine}. These food consumption indicators are collected with household surveys carried out by national institutions, or international organizations such as the WFP, the FAO and WB. The data collected in these surveys are not always openly accessible. The WFP provides daily estimations of the prevalence insufficient FCS and crisis or above rCSI obtained by phone surveys (the mobile Vulnerability Analysis and Monitoring, mVAM)  combined with in-person surveying through instruments like  the Comprehensive Food Security and Vulnerability Analysis (CFSVA) and the  Emergency Food Security Assessment (EFSA) (see Table 2 in \cite{machefer2024reviewML4FS}).  Previous studies have compiled several years of WFP population prevalence of insufficient food consumption indicators at ADMIN1 level\cite{martini2022machine, foini2023forecastability}. In the HFID, from the WFP-LIT source, we only include the face to face records as they are the only ones validated. The WFP-RT only contains phone surveys records (mVAM), collected using the remote modality, hence the separation of the variables in our HFID dataset (FCS-LIT, rCSI-LIT and FCS-RT, rCSI-RT).

The GADM \cite{gadm} dataset, composed of worldwide information on administrative boundaries at country and lower levels of sub-division, was used to have a common standardised reference for sub-national administrative units worldwide. In particular, we aimed at harmonizing the above mentioned food insecurity raw data sources into a dataset of monthly entries at GADM administrative level 2. 

All raw sources contain information on the location of the records in the form of (1) shape files with geometries or (2) with administrative name of each area concerned.  However different reference units are used (e.g. livelihood zones in the case of FEWS NET analyses). To harmonise geographies, we intersected the original geometries with the GADM areas at either ADMIN1 (for WFP population prevalences) or ADMIN2 granularity (for FEWS NET and IPC/CH analyses). Then each record was assigned to the GADM area with the largest intersection, provided that the sum of all intersections was covering more than $10\%$ of the original area \footnote{In case of geometries from raw sources with smaller units of analysis than the GADM ADMIN2, a potential artefact could be expected from this spatial-based approximation approach. On this  point, special attention is drawn on Haiti and Somalia from IPC/CH source and countries including very small livelihood zones in from FEWSNET source.}. WFP population prevalences were then repeated for all ADMIN2 contained in a given ADMIN1. For WFP-LIT, due to the lack of geo-referenced metadata, a pre-processing step to obtain geometries from administrative names was necessary to implement by using GPT$4.0$ together with Open Street Map (OSM). Geometries were assigned to the original locations (i.e. 'ADMIN0, ADMIN1' pairs) by using $osmnx$ (a python package based on OSM) and, when it was not possible to identify a given unit, we first corrected any misspelling or uncommon names by using the OpenAI API for GPT$4.0$. 

In order to build monthly entries, IPC/CH phases were repeated each month during the validity in the analysis, while FEWS NET phases, available three to four times per year, were only considered valid for the reference month. For the daily available WFP-RT raw indicators, we computed several monthly statistics -- minimum, maximum, average --  and for the daily available WFP-LIT raw indicators we computed the monthly average.

For IPC/CH and FEWS NET entries, we also retain information on distributed humanitarian aid as a boolean when analyses stipulate. Refugees and settlements presence was included as a boolean when indicated as such in the IPC API as per IPC mapping protocol (see IPC Manual \cite{ipcmanual3}, page 78). It is important to note that displaced populations might be part of the population analysed by the IPC even when not visually represented or classified as separate units, and therefore not listed as such in the API.

\begin{figure*}[ht]
\centering
\includegraphics[width=\linewidth]{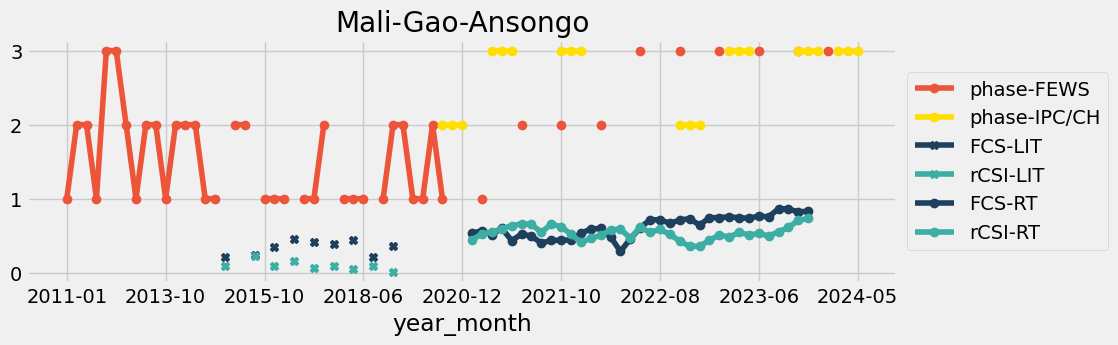}
\caption{This figure shows time series of  available variables values in the HFID (phase-FEWS, phase-IPC/CH, FCS-LIT, rCSI-LIT, FCS-RT, rCSI-RT) for the sub-region of Ansongo, Gao, Mali. A unique y-axis is used for display, reminding the reader that those variables represent different quantities.}
\label{fig:timeseries_variables}
\end{figure*}

\begin{figure}[ht]
    \centering
     \begin{subfigure}[b]{0.45\textwidth}
        \centering
        \includegraphics[width=\textwidth]{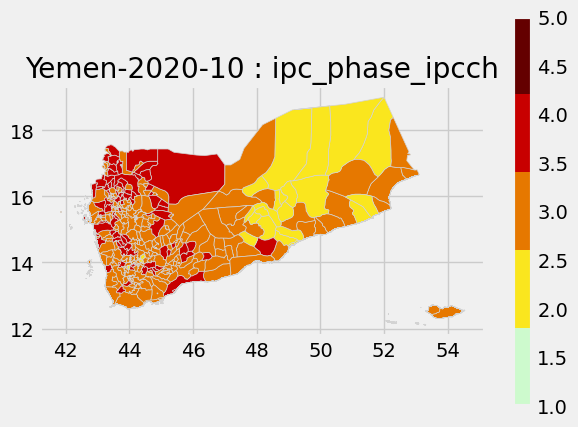}
        \caption{}
        \label{fig:map-ipcch}
    \end{subfigure}
    \hfill
    \begin{subfigure}[b]{0.45\textwidth}
        \centering
        \includegraphics[width=\textwidth]{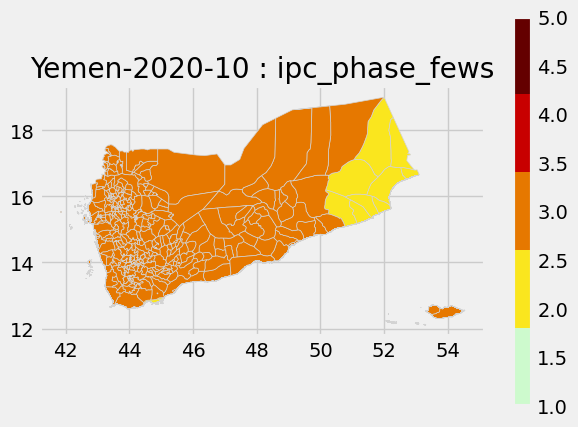}
        \caption{}
        \label{fig:map-fews}
    \end{subfigure}

    \vspace{0.5cm}

     \begin{subfigure}[b]{0.45\textwidth}
        \centering
        \includegraphics[width=\textwidth]{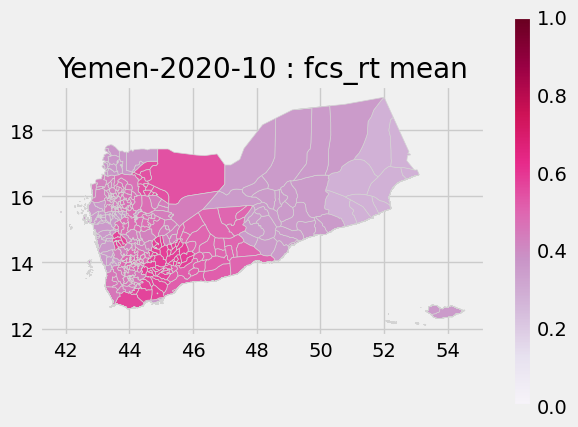}
        \caption{}
        \label{fig:map-fcs-rt}
    \end{subfigure}
    \hfill
    \begin{subfigure}[b]{0.45\textwidth}
        \centering
        \includegraphics[width=\textwidth]{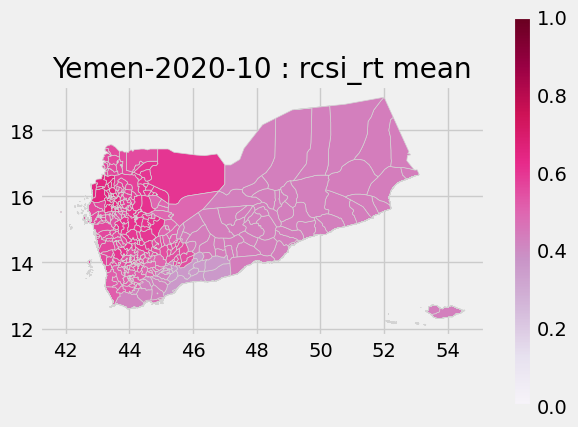}
        \caption{}
        \label{fig:map-rcsi-rt}
    \end{subfigure}
    
    \caption{An example of four mapped variables of the HFID for Yemen in October 2020: (a) phase-IPC/CH (variable \textit{ipc\_phase\_ipcch}), (b) phase-FEWS (variable \textit{ipc\_phase\_fews}), (d) monthly average of FCS-RT (variable \textit{fcs\_rt mean}), (e) monthly average of rCSI-RT (variable \textit{rcsi\_rt mean}). }
    \label{fig:maps-hfid}
\end{figure}

Finally, we merge all separate sources into a unique set of available datepoints at ADMIN2 and/or ADMIN1 (when no ADMIN2 data is available, i.e. in countries where only WFP population prevalence are found) entries. In the final dataset, WFP data are thus repeated for all those ADMIN2 units present in a given ADMIN1 when we have also records from IPC/CH or FEWS NET. An illustration of several variables timeseries of the HFID plotted for the ADMIN2 of Ansongo, located in Gao, Mali, is found in Fig.~\ref{fig:timeseries_variables}. An example of four HFID indices for Yemen on October $2020$ is given in Fig.~\ref{fig:maps-hfid}.

\section*{Data Records}
The HFID dataset is available on \textit{Scientific Data figshare currently only accessible to reviewers - will be replaced by an open access dataset link at publication}. The dataset consists of two files: one with the geometries (GADM.zip) and one with the HFID variables (HFID\_hv1.csv)

The GADM.zip file contains the shapefiles with administrative level 0 (gadm\_410\_L0),  level 1 (gadm\_410\_L1) and administrative level 2 (gadm\_410\_L2) reference geometries. 

The data table with the HFID variables (\textit{HFID\_hv1.csv}) is composed of several columns:
\begin{itemize}
    \item \textit{year\_month}: indicates the year and month of the record in the format "\%Y-\%m".
    \item \textit{ADMIN\_0}: name of the administrative level 0 (country) as defined in the GADM.
    \item \textit{ADMIN\_1}: name of the administrative level 1 (region) as defined in the GADM.
    \item \textit{ADMIN\_2}: name of the administrative level 2 (sub-region) as defined in the GADM.
     \item \textit{ipc\_phase\_fews}: FEWS NET-IPC compatible phase classification (phase-FEWS variable) if available, else NaN. Possible values: 1,2,3,4,5.
     \item  \textit{ha\_fews}: 1 if humanitarian aid has been distributed (owing to FEWS NET source), else NaN.
      \item \textit{ipc\_phase\_ipcch}: IPC/CH Phase classification (phase-IPC/CH variable) if available, else NaN. Possible values: 1,2,3,4,5,6.
     \item  \textit{ha\_ipcch}: 1 if humanitarian aid has been distributed (owing to IPC API source), else NaN.
     \item  \textit{set\_ipcch}: 1 if HouseHold Group or Internally Displaced People or  Urban settlements are found in the area (owing to IPC API source), else NaN.
     \item  \textit{rfg\_ipcch}: 1 if refugees are found in the area (owing to IPC API source), else NaN.
     \item \textit{fcs\_lit}: monthly average of  population prevalence of insufficient food consumption score from the WFP-LIT source (FCS-LIT variable) if available, else NaN. Possible values: between 0 and 1.
     \item \textit{rcsi\_lit} monthly average of  population prevalence of crisis or above reduced Coping Strategy Index from the WFP-LIT source (FCS-LIT variable) if available, else NaN. Possible values: between 0 and 1.
     \item \textit{fcs\_rt mean}:  monthly average of  population prevalence of insufficient food consumption score from the WFP-RT source (FCS-RT variable) if available, else NaN. Possible values: between 0 and 1.
     \item \textit{fcs\_rt max}:  monthly maximum of  population prevalence of insufficient food consumption score from the WFP-RT sourceif available, else NaN. Possible values: between 0 and 1. 
     \item \textit{fcs\_rt min}:  monthly minimum of  population prevalence of insufficient food consumption score from the WFP-RT source if available, else NaN. Possible values: between 0 and 1.
    \item \textit{rcsi\_rt mean}: monthly average of population prevalence of crisis or above reduced Coping Strategy Index from the WFP-RT source (FCS-RT variable) if available, else NaN. Possible values: between 0 and 1.
    \item \textit{rcsi\_rt max}: monthly maximum of population prevalence of crisis or above reduced Coping Strategy Index from the WFP-RT source if available, else NaN. Possible values: between 0 and 1.
    \item \textit{rcsi\_rt min}: monthly minimum of population prevalence of crisis or above reduced Coping Strategy Index from the WFP-RT source if available, else NaN. Possible values: between 0 and 1. 
    \item \textit{iso2}: ISO 3166-1 code for country in two letters.
    \item \textit{iso3}:  ISO 3166-1 code for country in three letters.
    \item \textit{region}: region name from the United Nations Geoscheme. 
\end{itemize}

\section*{Technical Validation}

One of the primary objectives of the HFID is to facilitate direct comparisons, at a single place, of various food insecurity variables across different areas and time periods, highlighting data availability and gaps. This is crucial for gaining a comprehensive understanding of regional conditions and enhancing the comparability of adopted food insecurity metrics.

\newcommand{\cellcolorADMINz}[1]{
    \ifnum#1>40
        \cellcolor[HTML]{006B3D}#1
    \else\ifnum#1>30
        \cellcolor[HTML]{069C56}#1
    \else\ifnum#1>20
        \cellcolor[HTML]{FF980E}#1
    \else\ifnum#1>10
        \cellcolor[HTML]{FF681E}#1
    \else
        \cellcolor[HTML]{D3212C}#1
    \fi\fi\fi\fi
}

\newcommand{\cellcolorADMINo}[1]{
    \ifnum#1>800
        \cellcolor[HTML]{006B3D}#1
    \else\ifnum#1>600
        \cellcolor[HTML]{069C56}#1
    \else\ifnum#1>400
        \cellcolor[HTML]{FF980E}#1
    \else\ifnum#1>200
        \cellcolor[HTML]{FF681E}#1
    \else
        \cellcolor[HTML]{D3212C}#1
    \fi\fi\fi\fi
}

\newcommand{\cellcolorADMINt}[1]{
    \ifnum#1>4500
        \cellcolor[HTML]{006B3D}#1
    \else\ifnum#1>3500
        \cellcolor[HTML]{069C56}#1
    \else\ifnum#1>2500
        \cellcolor[HTML]{FF980E}#1
    \else\ifnum#1>1500
        \cellcolor[HTML]{FF681E}#1
    \else
        \cellcolor[HTML]{D3212C}#1
    \fi\fi\fi\fi
}

\newcommand{\cellcolordatepoints}[1]{
    \ifnum#1>100
        \cellcolor[HTML]{006B3D}#1
    \else\ifnum#1>80
        \cellcolor[HTML]{069C56}#1
    \else\ifnum#1>60
        \cellcolor[HTML]{FF980E}#1
    \else\ifnum#1>40
        \cellcolor[HTML]{FF681E}#1
    \else
        \cellcolor[HTML]{D3212C}#1
    \fi\fi\fi\fi
}

\begin{table}[ht]
\centering
\begin{tabular}{l|ccccc}
\hline
& \textbf{ADMIN0} & \textbf{ADMIN1} & \textbf{ADMIN2} & \textbf{year-month} \\ \hline
\textbf{FCS-RT}  & \cellcolorADMINz{33} & \cellcolorADMINo{494} & \cellcolorADMINt{3301} & \cellcolordatepoints{67} \\
\textbf{rCSI-RT} & \cellcolorADMINz{33} & \cellcolorADMINo{494} & \cellcolorADMINt{3301} & \cellcolordatepoints{67} \\
\textbf{FCS-LIT}   & \cellcolorADMINz{70} & \cellcolorADMINo{978} & \cellcolorADMINt{3680} & \cellcolordatepoints{106}\\
\textbf{rCSI-LIT}  & \cellcolorADMINz{11} & \cellcolorADMINo{143} & \cellcolorADMINt{1434} & \cellcolordatepoints{46} \\
\textbf{phase-FEWS}   & \cellcolorADMINz{28} & \cellcolorADMINo{483} & \cellcolorADMINt{3911} & \cellcolordatepoints{46} \\
\textbf{phase-IPC/CH}    & \cellcolorADMINz{49} & \cellcolorADMINo{678} & \cellcolorADMINt{5022} & \cellcolordatepoints{89}  \\ \hline

\end{tabular}
\caption{\label{tab:stats-admins} Number of unique countries (ADMIN0), and administrative level 1 (ADMIN1) and 2 (ADMIN2) areas, year-month entries for each variable of the HFID.}
\end{table}

\newcommand{\cellcolorbyvalue}[1]{
    \ifnum#1>100000
        \cellcolor[HTML]{006B3D}#1
    \else\ifnum#1>50000
        \cellcolor[HTML]{069C56}#1
    \else\ifnum#1>10000
        \cellcolor[HTML]{FF980E}#1
    \else\ifnum#1>1000
        \cellcolor[HTML]{FF681E}#1
    \else
        \cellcolor[HTML]{D3212C}#1
    \fi\fi\fi\fi
}
\begin{table}[ht]
\centering
\begin{tabular}{l|rrrrrr}
\hline
                  & \textbf{FCS-RT} & \textbf{rCSI-RT} & \textbf{FCS -LIT} & \textbf{rCSI-LIT} & \textbf{phase-FEWS} & \textbf{phase-IPC/CH} \\ \hline
\textbf{FCS-RT}  & \cellcolorbyvalue{122112} & & & & & \\
\textbf{rCSI-RT} & \cellcolorbyvalue{122112} & \cellcolorbyvalue{122112} & & & & \\
\textbf{FCS-LIT}   & \cellcolorbyvalue{0} & \cellcolorbyvalue{0} & \cellcolorbyvalue{19078} & & & \\
\textbf{rCSI-LIT}  & \cellcolorbyvalue{0} & \cellcolorbyvalue{0} & \cellcolorbyvalue{12583} & \cellcolorbyvalue{15961} & & \\
\textbf{phase-FEWS}   & \cellcolorbyvalue{23662} & \cellcolorbyvalue{23662} & \cellcolorbyvalue{3443} & \cellcolorbyvalue{3666} & \cellcolorbyvalue{138334} & \\
\textbf{phase-IPC/CH}    & \cellcolorbyvalue{32433} & \cellcolorbyvalue{32433} & \cellcolorbyvalue{719} & \cellcolorbyvalue{17} & \cellcolorbyvalue{12618} & \cellcolorbyvalue{97478} \\ \hline
\end{tabular}
\caption{\label{tab:stats-records} Number of matching records between variables of the HFID.}
\end{table}

From Tab.~\ref{tab:stats-admins} and ~\ref{tab:stats-records}, we find that the HFID has the broadest geographical and historical coverage for the phase-IPC/CH and FCS-LIT variables, with around $100,000$ records in respectively $49$ and $70$ countries over $89$ and $106$ year-month entries and a high number of ADMIN2 covered. However, we highlight that FCS-LIT variable is only available until July 2021.

\begin{figure*}[ht]
\centering
\includegraphics[width=\linewidth]{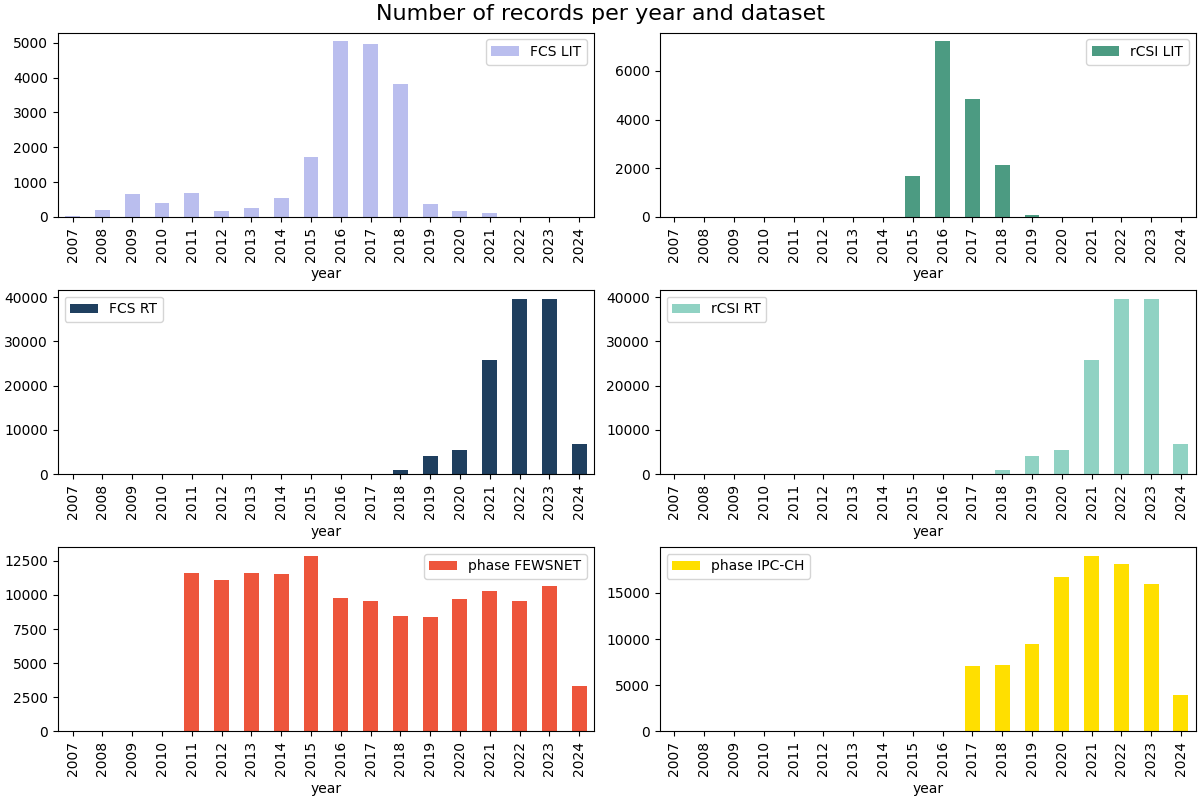}
\caption{Number of records per year and variable of the HFID.}
\label{fig:timeline_year}
\end{figure*}

We therefore subsequently analyse the temporal availability of the different variables of the HFID dataset in Fig.\ref{fig:timeline_year}, plotting the number of records per year. This presentation facilitates the understanding of data gaps. We observe that the WFP-LIT provides a peak of records in 2016, and other still ongoing sources (WFP-RT, phase-FEWS and phase-IPC/CH) offer quite stable number of records since 2022 at least (2024 is still ongoing year, thus with less records). We observe that for WFP-LIT, pre-2015 records are very sparse. The variable phase-FEWS provides the longest time serie and most steady and numerous records (since 2011, over $10,000$ yearly records), and the variable phase-IPC/CH has been ramping up from around $7,000$ records in 2017 up to more than $15,000$ since 2020.

\begin{figure*}[ht]
\centering
\includegraphics[width=\linewidth]{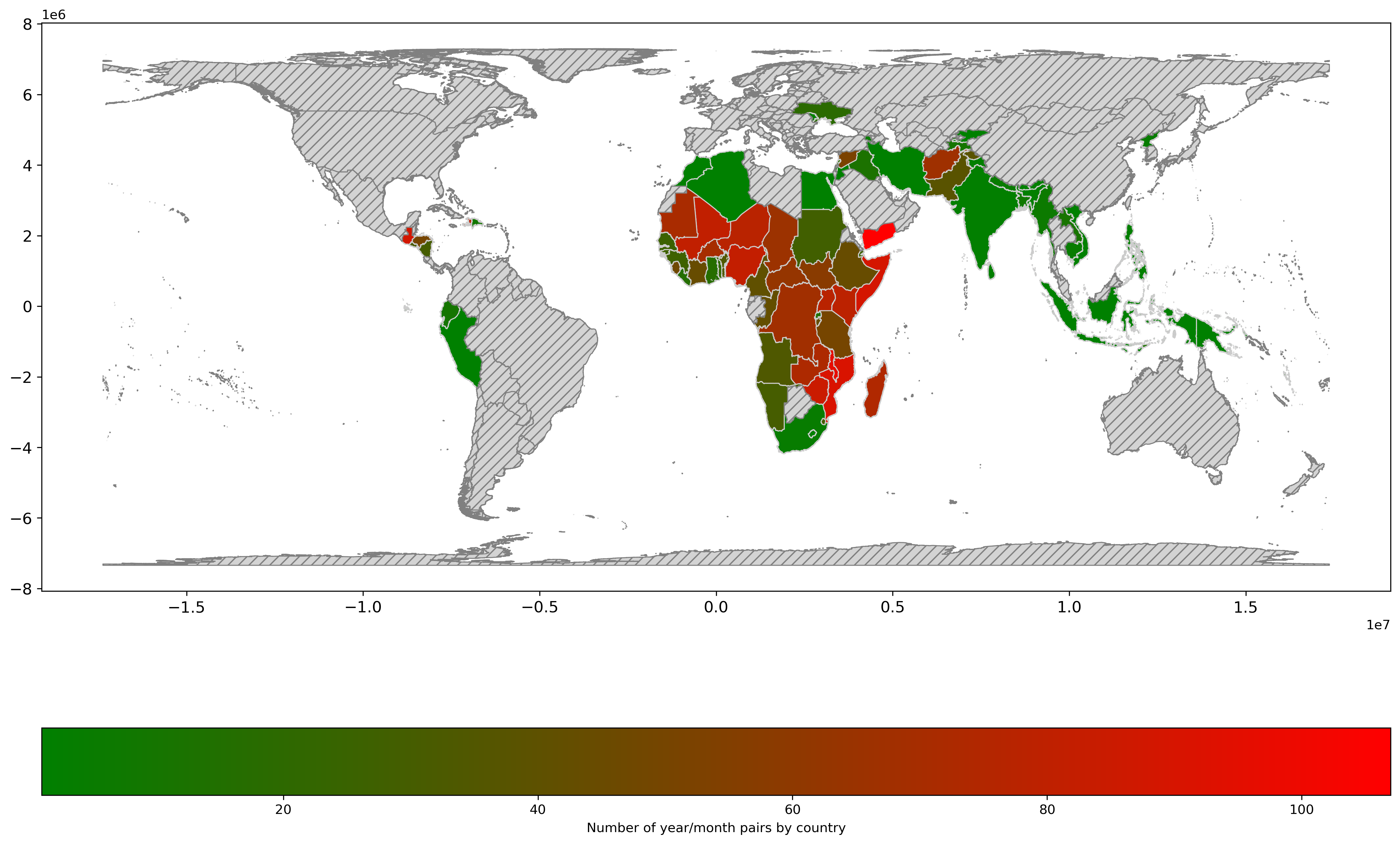}
\caption{The world map coverage of HFID dataset in terms of unique \textit{year\_month} occurrences by ADMIN0 (country) for all variables. }
\label{fig:map-yearmonth}
\end{figure*}

\begin{figure*}[ht]
\centering
\includegraphics[width=\linewidth]{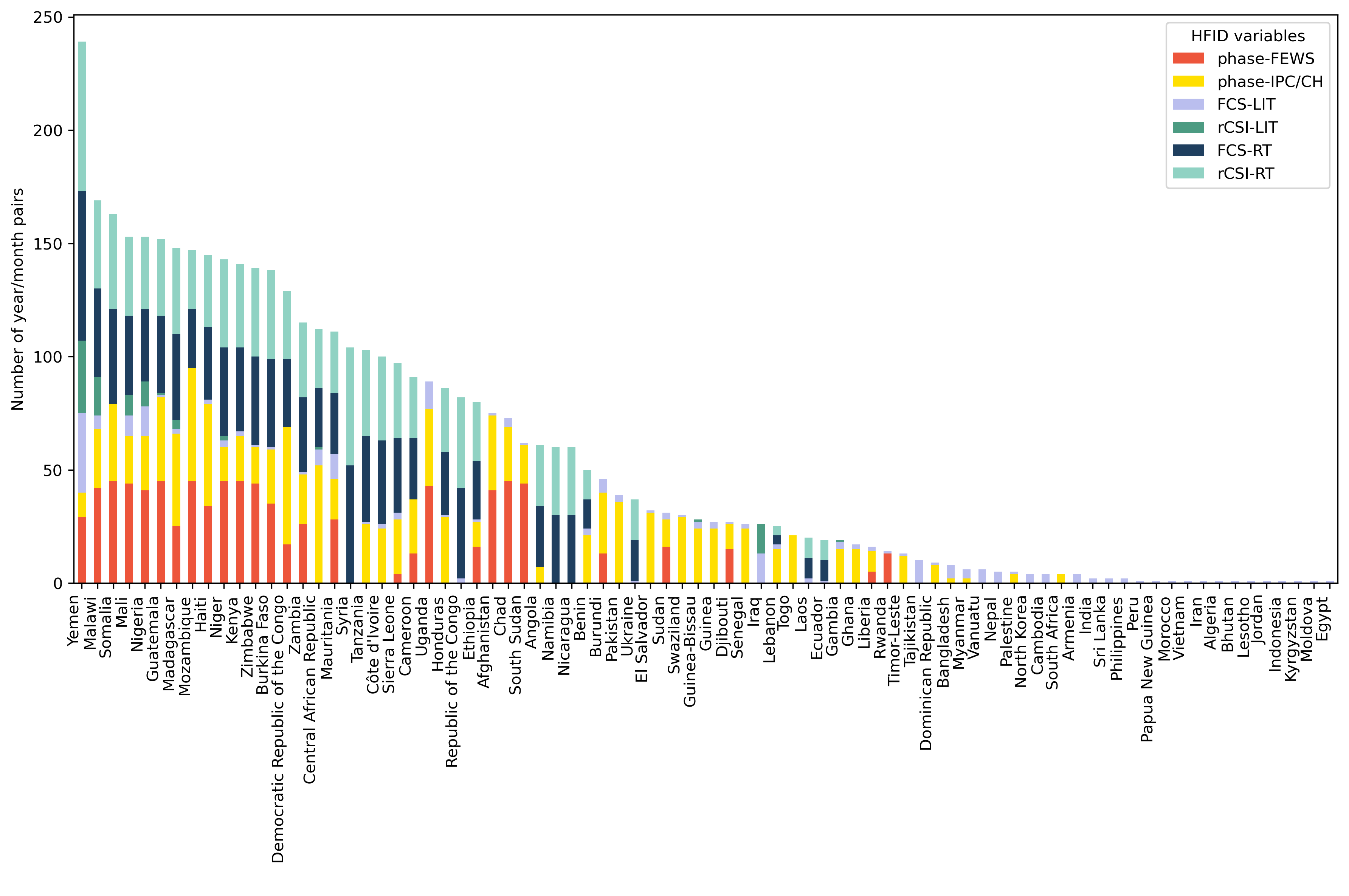}
\caption{Number of unique year/month pairs by variable of the HFID dataset and by ADMIN0 (country).}
\label{fig:year_month_histo_adm1}
\end{figure*}

We also analyse the number of unique \textit{year\_month} occurrences by country (ADMIN0) in Fig.~\ref{fig:map-yearmonth}. The detail by HFID variable and country is provided in Fig.~\ref{fig:year_month_histo_adm1}. More data both in terms of years and number of data sources is available for Latin America and the Caribbean, Western Africa, Middle Africa, Eastern Africa, Western Asia, and Southern Asia. On the contrary Eastern Europe, Central Asia, Eastern Asia, South-Eastern Asia and Oceania have very limited coverage in terms of the number of HFID variables as well as number of years \footnote{The regions we use to described the data are from the United Nations Geoscheme.}. These illustrations also reveal that Yemen, the  Sahelian countries Mali, Niger, Nigeria as well as Somalia, Malawi, Madagascar and Mozambique in Eastern Africa  have the  largest number of unique \textit{year\_month} occurrences.

\begin{figure*}[ht]
\centering
\includegraphics[width=\linewidth]{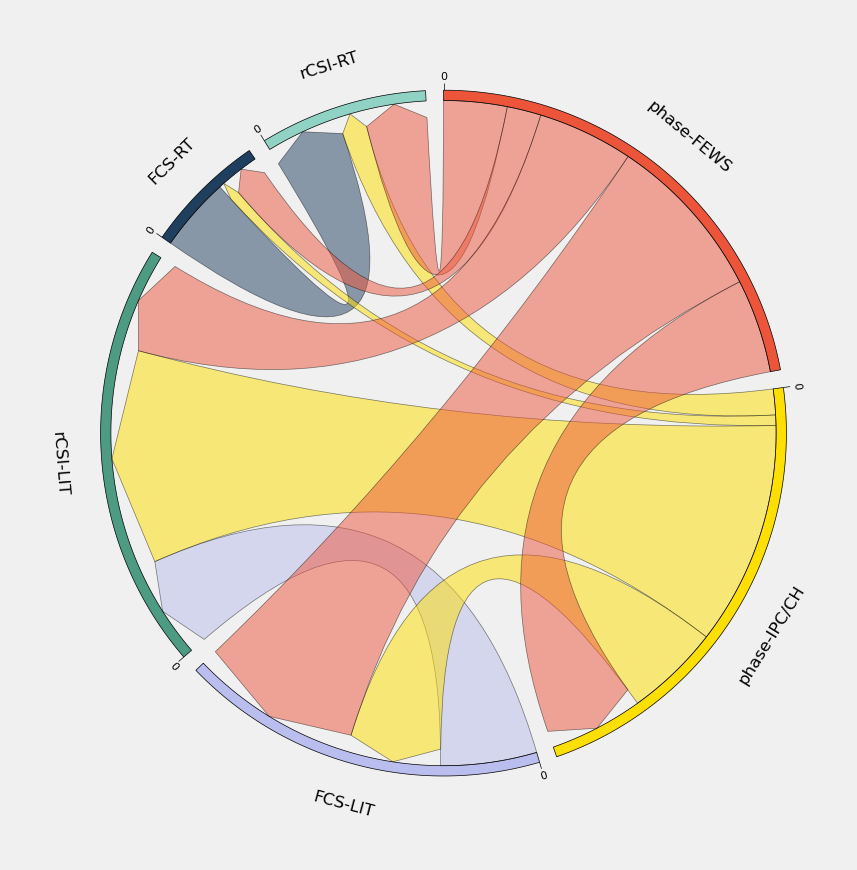}
\caption{Chord plot showing the relation among all the variables in the HFID. The size of the arrows is proportional to the mutual information score among each pair of indicators. This provides an indication on the amount of information gained about one indicator through the observation of a second indicator in the HFID.}
\label{fig:chordpl}
\end{figure*}

In order to compare values of the variables of the HFID, a basic statistical analysis consists in computing the correlation between variables, as found in Tab.~\ref{tab:correlation}. Some degree of correlation might exist due to the definition of some of the variables of the HFID, although not necessarily, noteworthy the different nature of the variables do not measure the same aspects of the dimension of food security \cite{machefer2024reviewML4FS}. We find that the FCS-LIT strongly correlates with all the other variables, with weakest correlation obtained for phase-IPC/CH which is also the variable for which there is however the least matching records (see column FCS-LIT in Tab.~\ref{tab:stats-records}). Conversely, the phase-IPC/CH variable very weakly correlate with the FCS-RT while there is  for a reasonable amount of records having both variables ($32,433$ see Tab.~\ref{tab:stats-records}). Weak correlation is also found for WFP-RT variables with phase-FEWS and phase-IPC/CH variables. Strong correlation is observed between phase-FEWS and phase-IPC/CH variables. Dependencies can also be non-linear and, thus, we also provide a comparison among all HFID indices in terms of mutual information score in Fig.~\ref{fig:chordpl}. Mutual information is a way to measure how much one piece of data tells us about another. It captures both linear and nonlinear dependencies between variables, surpassing linear correlation measures in identifying complex relationships\cite{infoth}. To visually represent this, we can utilize arrows: larger arrows indicate stronger connections, and it is important to note that the directions are arbitrary and can be interpreted in both ways. According to Fig.~\ref{fig:chordpl}, WFP-LIT variables carry more information on the phases than WFP-RT data.

\begin{table}[ht]
\centering
\begin{tabular}{l|cccccc}
\hline
                  & \textbf{FCS-RT} & \textbf{rCSI-RT} & \textbf{FCS-LIT} & \textbf{rCSI-LIT} & \textbf{phase-FEWS} & \textbf{phase-IPC/CH} \\ \hline
\textbf{FCS-RT}  & \cellcolor{green!99!red!10} 1.00 &  &  &  &  &  \\ 
\textbf{rCSI-RT} & \cellcolor{green!39!red!61} 0.39 & \cellcolor{green!99!red!10} 1.00 &  &  &  &  \\ 
\textbf{FCS-LIT}   &  &  & \cellcolor{green!99!red!10} 1.00 &  &  &  \\ 
\textbf{rCSI-LIT}  &  &  & \cellcolor{green!65!red!35} 0.65 & \cellcolor{green!99!red!10} 1.00 &  &  \\ 
\textbf{phase-FEWS}   & \cellcolor{green!32!red!68} 0.32 & \cellcolor{green!49!red!51} 0.49 & \cellcolor{green!73!red!27} 0.73 & \cellcolor{green!45!red!55} 0.45 & \cellcolor{green!99!red!10} 1.00 &  \\ 
\textbf{phase-IPC/CH}    & \cellcolor{green!15!red!95} 0.02 & \cellcolor{green!30!red!70} 0.30 & \cellcolor{green!53!red!47} 0.53 & \cellcolor{green!39!red!61} 0.39 & \cellcolor{green!61!red!39} 0.61 & \cellcolor{green!99!red!10} 1.00 \\ \hline

\end{tabular}
\caption{\label{tab:correlation} Correlation of HFID variables. }
\end{table}

\begin{figure*}[ht]
\centering
\includegraphics[width=\linewidth]{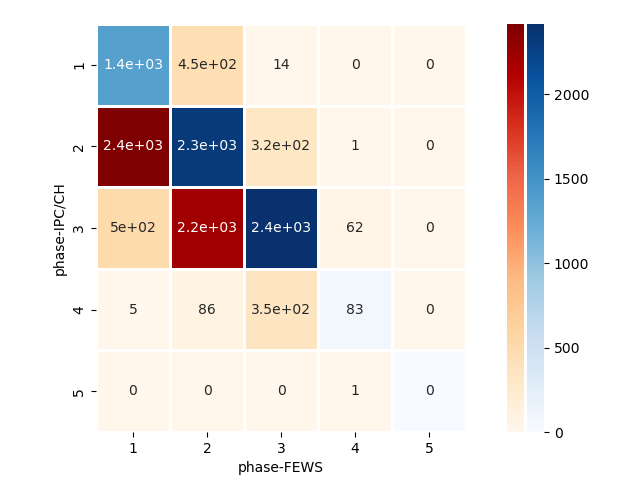}
\caption{The figure illustrates the number of records in agreement between FEWS NET phases (phase-FEWS) in x-axis and IPC/CH phases (phase-IPC/CH) in y-axis. In an ideal scenario of complete agreement between the two classification schemes, the heatmap would form a diagonal matrix. Off-diagonal elements provide insight into areas of disagreement between the two schemes.}
\label{fig:hmap}
\end{figure*}

We compare the phase-FEWS and phase-IPC/CH values in Fig.~\ref{fig:hmap}. The HFID contains $12,618$ records with both sources, covering $22$ countries. It is worth noting that while there is some general consensus between the two classification values, phase-FEWS consistently trend lower than phase-IPC/CH. Specifically, when the phase-IPC/CH is equal to 2, the phase-FEWS is observed to be either 1 or 2 with the same frequency. The same pattern is observed for phase-IPC equal to 3. As far as we know, this is the first comprehensive comparison of these two classification schemes.

\begin{figure*}[ht]
\centering
\includegraphics[width=\linewidth]{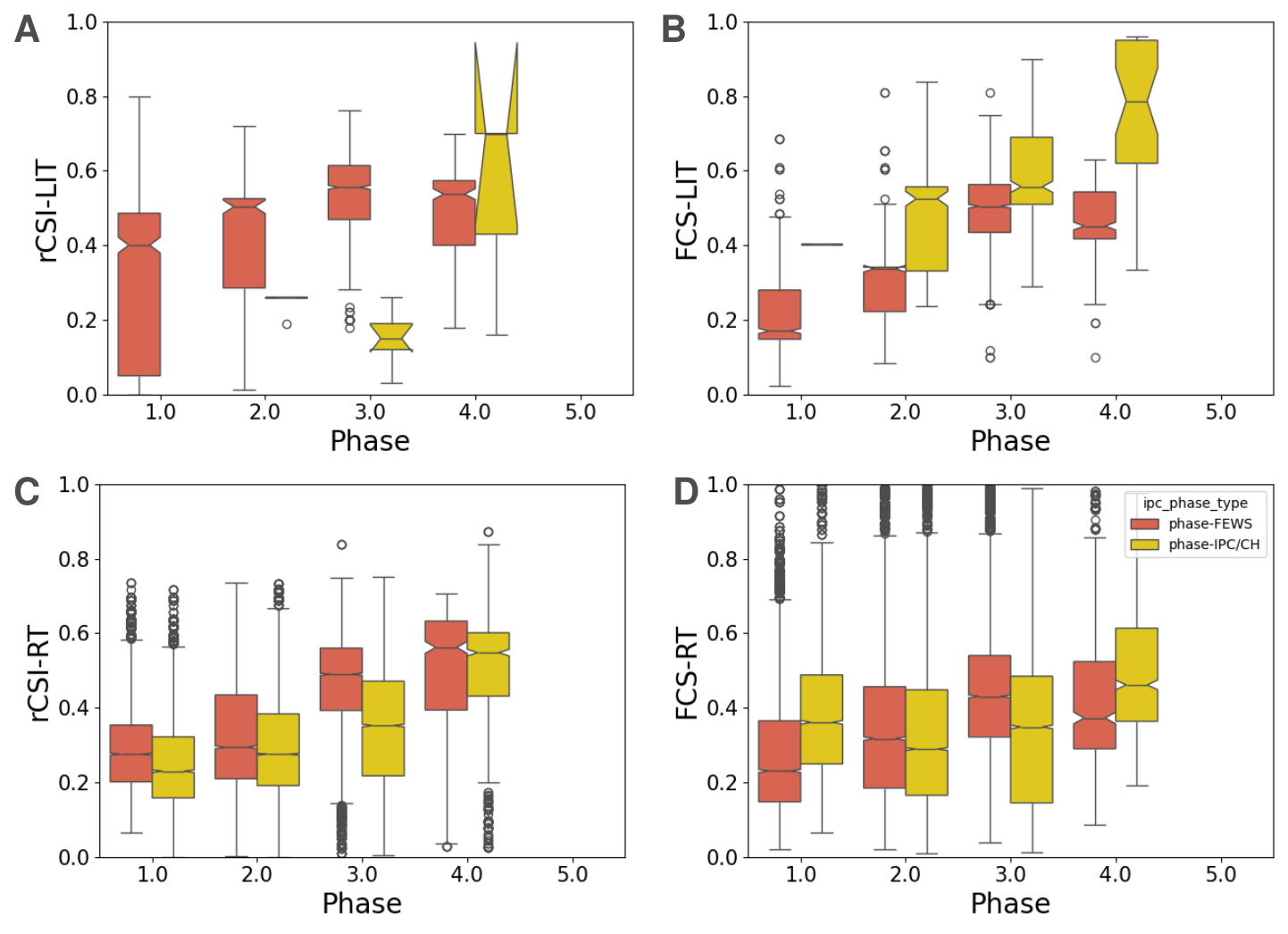}
\caption{Box plots representing the distribution population prevalence of insufficient food consumption indicators (FCS, rCSI) from WFP-LIT and from WFP-RT, segmented into five different classes, distinguishing between FEWS NET and IPC/CH classifications.}
\label{fig:boxpl}
\end{figure*}

In Fig.~\ref{fig:boxpl} we present a comparison of the FCS-LIT, FCS-RT, rCSI-LIT and rCSI-RT\footnote{For WFP-RT source, we take the monthly averaged variables \textit{fcs\_rt mean} and \textit{fcs\_rt mean} (see Section Data Records)} over all administrative units, both among themselves and in relation to both phase-IPC/CH and phase-FEWS variables, when spatio-temporal matching records are found. We show the results for the five phases provided by either FEWS NET(red boxes) or IPC/CH (yellow boxes). The number of records varies depending on the availability of pairs of indices for the same unit and year/month pair. As expected, we observe an increase in population prevalence of insufficient FCS and crisis or above rCSI across all phases. However, there are also differences between WFP-LIT and WFP-RT sources. Both rCSI-LIT ($3,666$ events) and FCS-LIT ($3,443$ events) saturate for crisis levels of phase-FEWS (i.e. $\geq 3$), while the (median) prevalence keeps increasing almost linearly over phase-IPC/CH (take into account the very limited statistics, having only $17$ events for rCSI-LIT and $719$ for FCS-LIT). Conversely, WFP-RT data exhibits a more uniform pattern, indicating a similar trend for both classification schemes. Specifically, the rCSI-RT consistently increases for higher phase-FEWS ($23,662$ events) and phase-IPC/CH ($32,433$ events) classes, as shown in the bottom panel. FCS-RT also shows an increase in correlation with phase-FEWS ($23,662$ events) and phase-IPC/CH ($32,433$ events) values, but it is not as responsive to changes in phase (see also Tab.~\ref{tab:correlation}). This also indicates that rCSI-RT may be a more accurate predictor than FCS-RT for both FEWS NET and IPC/CH phase variables. However, it should be kept in mind that phase-IPC/CH 4 or more, FCS and rCSI are non defining characteristics in the IPC/CH protocol (see the IPC Acute Food Insecurity Reference Table\cite{ipcmanual3}, page 37), meaning that analysts need to use other indicators (such as the Household Hunger Scale or Household Economy Scale) to discriminate between phase 4 and 5.

\section*{Usage Notes}

The code is reproducible and modulable in order to add future records, on a monthly or yearly basis, from existing sources (FEWS NET, IPC/CH, WFP-RT) and potential other sources of interest, like the population prevalence of answers to surveys enabling single food security indicator computation from the \href{https://data-in-emergencies.fao.org/search?groupIds=ab8a43038b6347ac93507988f7e2a90b&q=Food%20security%20Thematic%20Area}{FAO Data in Emergencies}. The HFID, structured in a tabular format, facilitates its integration into data-driven and machine learning analyses. Researchers and analysts can conveniently aggregate selected covariates at various levels of granularity, ranging from admin 2 to country level, suiting the specific needs of their studies. This adaptability makes the HFID an instrumental dataset for advanced analytical approaches in food security.

We also provide with some functions automatically reading and processing the HFID dataset, in module \textit{targets.utils.utils.py}:
\begin{itemize}
    \item the function \textit{read\_hfid()} reads the HFID tabular dataset (HFID\_hv1.csv) enabling the consideration that the "iso2" column with the "NA" value represents "Namibia" country and not a NaN.
    \item the function \textit{read\_admins()} reads all the shapefiles with specified administrative level reference geometries (GADM.zip). The processing includes the suppression of specific sub regions where inter-national conflicts of identity exist (e.g. China, India, Pakistan, Kosovo).
    \item the function \textit{sanitize\_high\_phases()} allows to decide on the treatment to apply to IPC/CH Phase 6 (keep as such, remove it, or cap it to Phase 5).
    \item the function \textit{add\_admins2\_geometry\_to\_hfid()} which combines the HFID tabular dataset and the administrative level geometries.
\end{itemize}

Finally, we propose some vizualisation functions, enabling to plot maps of the variables of the HFID dataset, in module \textit{targets.utils.plot\_utils.py}:
\begin{itemize}
 \item the function \textit{plot\_phases\_map()} plots a map with phases from FEWS NET or IPC/CH (variables phase-FEWS, phase-IPC/CH) at a certain date (\textit{year\_month}), respecting the IPC color code. An illustration of the output of this function is found in  figures Fig.~\ref{fig:map-ipcch} and Fig.~\ref{fig:map-fews}.
    \item the function \textit{plot\_outcomes\_indc\_map()} plots a map with values from population prevalence of insufficient food consumption indicators from WFP-LIT and WFP-RT (variables FCS-LIT, rCSI-LIT, FCS-RT, rCSI-RT) at a certain date (\textit{year\_month}). An illustration of the output of this function is found in figures Fig.~\ref{fig:map-fcs-rt} and Fig.~\ref{fig:map-rcsi-rt}.
\end{itemize}

\section*{Code availability}

The code and relevant documentation is available on \textit{Scientific Data zipfile only accessible to reviewers (will be replaced by an open access link at publication)} and is encouraged to be used citing our work in publications and reports, and using our license in further published codebase by external users.

\bibliography{sample}

\begin{thebibliography}{10}
\urlstyle{rm}
\expandafter\ifx\csname url\endcsname\relax
  \def\url#1{\texttt{#1}}\fi
\expandafter\ifx\csname urlprefix\endcsname\relax\def\urlprefix{URL }\fi
\expandafter\ifx\csname doiprefix\endcsname\relax\def\doiprefix{DOI: }\fi
\providecommand{\bibinfo}[2]{#2}
\providecommand{\eprint}[2][]{\url{#2}}

\bibitem{SDG23}
\bibinfo{author}{UN}.
\newblock \bibinfo{title}{The sustainable development goals report 2023}.
\newblock \bibinfo{howpublished}{Available at \url{https://unstats.un.org/sdgs/report/2023/The-Sustainable-Development-Goals-Report-2023.pdf}} (\bibinfo{year}{2023}).

\bibitem{machefer2024reviewML4FS}
\bibinfo{author}{Machefer, M.} \emph{et~al.}
\newblock \bibinfo{journal}{\bibinfo{title}{Potential and limitations of machine learning modelling for forecasting acute food insecurity}}.
\newblock {\emph{\JournalTitle{arXiv preprint under preparation}}}  (\bibinfo{year}{2024}).

\bibitem{cafiero2014}
\bibinfo{author}{Cafiero, C.}, \bibinfo{author}{Melgar-Quiñonez, H.~R.}, \bibinfo{author}{Ballard, T.~J.} \& \bibinfo{author}{Kepple, A.~W.}
\newblock \bibinfo{journal}{\bibinfo{title}{Validity and reliability of food security measures}}.
\newblock {\emph{\JournalTitle{Annals of the New York Academy of Sciences}}} \textbf{\bibinfo{volume}{1331}}, \bibinfo{pages}{230--248} (\bibinfo{year}{2014}).

\bibitem{Maxwell2008}
\bibinfo{author}{Maxwell, D.}, \bibinfo{author}{Caldwell, R.} \& \bibinfo{author}{Langworthy, M.}
\newblock \bibinfo{journal}{\bibinfo{title}{Measuring food insecurity: Can an indicator based on localized coping behaviors be used to compare across contexts?}}
\newblock {\emph{\JournalTitle{Food Policy}}} \textbf{\bibinfo{volume}{33}}, \bibinfo{pages}{533--540}, \url{https://doi.org/10.1016/j.foodpol.2008.02.004} (\bibinfo{year}{2008}).

\bibitem{Coates2013}
\bibinfo{author}{Coates, J.}
\newblock \bibinfo{journal}{\bibinfo{title}{Build it back better: Deconstructing food security for improved measurement and action}}.
\newblock {\emph{\JournalTitle{Global Food Security}}} \textbf{\bibinfo{volume}{2}}, \bibinfo{pages}{188--194}, \url{https://doi.org/10.1016/j.gfs.2013.05.002} (\bibinfo{year}{2013}).

\bibitem{Headey2013}
\bibinfo{author}{Headey, D.} \& \bibinfo{author}{Ecker, O.}
\newblock \bibinfo{journal}{\bibinfo{title}{Rethinking the measurement of food security: from first principles to best practice}}.
\newblock {\emph{\JournalTitle{Food Sec}}} \textbf{\bibinfo{volume}{5}}, \bibinfo{pages}{327–343}, \url{https://doi.org/10.1007/s12571-013-0253-0} (\bibinfo{year}{2013}).

\bibitem{Maxwell2014}
\bibinfo{author}{Maxwell, D.}, \bibinfo{author}{Vaitla, B.} \& \bibinfo{author}{Coates, J.}
\newblock \bibinfo{journal}{\bibinfo{title}{How do indicators of household food insecurity measure up? an empirical comparison from ethiopia}}.
\newblock {\emph{\JournalTitle{Food Policy}}} \textbf{\bibinfo{volume}{47}}, \bibinfo{pages}{107--116}, \url{https://doi.org/10.1016/j.foodpol.2014.04.003} (\bibinfo{year}{2014}).

\bibitem{Allee2021}
\bibinfo{author}{Allee, A.}, \bibinfo{author}{Lynd, L.} \& \bibinfo{author}{Vaze, V.}
\newblock \bibinfo{journal}{\bibinfo{title}{Cross-national analysis of food security drivers: comparing results based on the food insecurity experience scale and global food security index}}.
\newblock {\emph{\JournalTitle{Food Sec}}} \textbf{\bibinfo{volume}{13}}, \bibinfo{pages}{1245–1261}, \url{https://doi.org/10.1007/s12571-021-01156-w} (\bibinfo{year}{2013}).

\bibitem{lentz2019data}
\bibinfo{author}{Lentz, E.~C.}, \bibinfo{author}{Michelson, H.}, \bibinfo{author}{Baylis, K.} \& \bibinfo{author}{Zhou, Y.}
\newblock \bibinfo{journal}{\bibinfo{title}{A data-driven approach improves food insecurity crisis prediction}}.
\newblock {\emph{\JournalTitle{World Development}}} \textbf{\bibinfo{volume}{122}}, \bibinfo{pages}{399--409} (\bibinfo{year}{2019}).

\bibitem{andree2020predicting}
\bibinfo{author}{Andree, B. P.~J.}, \bibinfo{author}{Chamorro, A.}, \bibinfo{author}{Kraay, A.}, \bibinfo{author}{Spencer, P.} \& \bibinfo{author}{Wang, D.}
\newblock \emph{\bibinfo{title}{Predicting food crises}} (\bibinfo{publisher}{The World Bank}, \bibinfo{year}{2020}).

\bibitem{zhou2022machine}
\bibinfo{author}{Zhou, Y.}, \bibinfo{author}{Lentz, E.}, \bibinfo{author}{Michelson, H.}, \bibinfo{author}{Kim, C.} \& \bibinfo{author}{Baylis, K.}
\newblock \bibinfo{journal}{\bibinfo{title}{Machine learning for food security: Principles for transparency and usability}}.
\newblock {\emph{\JournalTitle{Applied Economic Perspectives and Policy}}} \textbf{\bibinfo{volume}{44}}, \bibinfo{pages}{893--910} (\bibinfo{year}{2022}).

\bibitem{westerveld2021forecasting}
\bibinfo{author}{Westerveld, J.~J.} \emph{et~al.}
\newblock \bibinfo{journal}{\bibinfo{title}{Forecasting transitions in the state of food security with machine learning using transferable features}}.
\newblock {\emph{\JournalTitle{Science of the Total Environment}}} \textbf{\bibinfo{volume}{786}}, \bibinfo{pages}{147366} (\bibinfo{year}{2021}).

\bibitem{deleglise2022food}
\bibinfo{author}{Del{\'e}glise, H.} \emph{et~al.}
\newblock \bibinfo{journal}{\bibinfo{title}{Food security prediction from heterogeneous data combining machine and deep learning methods}}.
\newblock {\emph{\JournalTitle{Expert Systems with Applications}}} \textbf{\bibinfo{volume}{190}}, \bibinfo{pages}{116189} (\bibinfo{year}{2022}).

\bibitem{wang2022transitions}
\bibinfo{author}{Wang, D.}, \bibinfo{author}{Andr{\'e}e, B. P.~J.}, \bibinfo{author}{Chamorro, A.~F.} \& \bibinfo{author}{Spencer, P.~G.}
\newblock \bibinfo{journal}{\bibinfo{title}{Transitions into and out of food insecurity: A probabilistic approach with panel data evidence from 15 countries}}.
\newblock {\emph{\JournalTitle{World Development}}} \textbf{\bibinfo{volume}{159}}, \bibinfo{pages}{106035} (\bibinfo{year}{2022}).

\bibitem{martini2022machine}
\bibinfo{author}{Martini, G.} \emph{et~al.}
\newblock \bibinfo{journal}{\bibinfo{title}{Machine learning can guide food security efforts when primary data are not available}}.
\newblock {\emph{\JournalTitle{Nature Food}}} \textbf{\bibinfo{volume}{3}}, \bibinfo{pages}{716--728} (\bibinfo{year}{2022}).

\bibitem{krishnamurthy2022anticipating}
\bibinfo{author}{Krishnamurthy~R, P.~K.}, \bibinfo{author}{Fisher, J.~B.}, \bibinfo{author}{Choularton, R.~J.} \& \bibinfo{author}{Kareiva, P.~M.}
\newblock \bibinfo{journal}{\bibinfo{title}{Anticipating drought-related food security changes}}.
\newblock {\emph{\JournalTitle{Nature Sustainability}}} \textbf{\bibinfo{volume}{5}}, \bibinfo{pages}{956--964} (\bibinfo{year}{2022}).

\bibitem{foini2023forecastability}
\bibinfo{author}{Foini, P.}, \bibinfo{author}{Tizzoni, M.}, \bibinfo{author}{Martini, G.}, \bibinfo{author}{Paolotti, D.} \& \bibinfo{author}{Omodei, E.}
\newblock \bibinfo{journal}{\bibinfo{title}{On the forecastability of food insecurity}}.
\newblock {\emph{\JournalTitle{Scientific Reports}}} \textbf{\bibinfo{volume}{13}}, \bibinfo{pages}{2793} (\bibinfo{year}{2023}).

\bibitem{busker2023predicting}
\bibinfo{author}{Busker, T.~S.} \emph{et~al.}
\newblock \bibinfo{journal}{\bibinfo{title}{Predicting food-security crises in the horn of africa using machine learning}}.
\newblock {\emph{\JournalTitle{Authorea Preprints}}}  (\bibinfo{year}{2023}).

\bibitem{herteux2023forecasting}
\bibinfo{author}{Herteux, J.}, \bibinfo{author}{R{\"a}th, C.}, \bibinfo{author}{Baha, A.}, \bibinfo{author}{Martini, G.} \& \bibinfo{author}{Piovani, D.}
\newblock \bibinfo{journal}{\bibinfo{title}{Forecasting trends in food security: a reservoir computing approach}}.
\newblock {\emph{\JournalTitle{arXiv preprint arXiv:2312.00626}}}  (\bibinfo{year}{2023}).

\bibitem{IPC}
\bibinfo{author}{IPC}.
\newblock \bibinfo{title}{The integrated food security phase classification}.
\newblock \bibinfo{howpublished}{Available at \url{https://www.ipcinfo.org/fileadmin/user_upload/ipcinfo/manual/IPC_Technical_Manual_3_Final.pdf}} (\bibinfo{year}{2021}).

\bibitem{FEWSNET}
\bibinfo{author}{FEWSNET}.
\newblock \bibinfo{title}{Famine early warning system network}.
\newblock \bibinfo{howpublished}{Available at \url{https://fews.net/}}.

\bibitem{WFP}
\bibinfo{author}{WFP}.
\newblock \bibinfo{title}{Food security analysis}.
\newblock \bibinfo{howpublished}{Available at \url{https://www.wfp.org/food-security-analysis}}.

\bibitem{gadm}
\bibinfo{author}{GADM}.
\newblock \bibinfo{title}{Gadm. database of global administrative areas}.
\newblock \bibinfo{howpublished}{Available at \url{https://gadm.org/data.html }} (\bibinfo{year}{2022}).

\bibitem{Caccavale2020}
\bibinfo{author}{Caccavale, O.~M.} \& \bibinfo{author}{Giuffrida, V.}
\newblock \bibinfo{journal}{\bibinfo{title}{The proteus composite index: Towards a better metric for global food security}}.
\newblock {\emph{\JournalTitle{World Development}}} \textbf{\bibinfo{volume}{126}}, \bibinfo{pages}{104709}, \url{https://doi.org/10.1016/j.worlddev.2019.104709} (\bibinfo{year}{2020}).

\bibitem{ipcaccuracystudy}
\bibinfo{author}{Lentz, E.}, \bibinfo{author}{Baylis, K.}, \bibinfo{author}{Michelson, H.} \& \bibinfo{author}{Kim, C.}
\newblock \bibinfo{title}{Ipc accuracy study: analyzing the internal consistency of ipc afi and amn analyses}.
\newblock \bibinfo{howpublished}{\url{https://www.ipcinfo.org/fileadmin/user_upload/ipcinfo/docs/IPC_Accuracy_Study.pdf}} (\bibinfo{year}{2024}).
\newblock \bibinfo{note}{Accessed: 2024-05-31}.

\bibitem{api_fews}
\bibinfo{author}{FEWSNET}.
\newblock \bibinfo{title}{Famine early warning system network api}.
\newblock \bibinfo{howpublished}{Available at \url{https://fdw.fews.net/api/ipcpackage/}}.

\bibitem{api_ipc}
\bibinfo{author}{IPC/CH}.
\newblock \bibinfo{title}{Integrated phase classification cadre harmonise api}.
\newblock \bibinfo{howpublished}{Available at \url{https://api.ipcinfo.org/analysis/}}.

\bibitem{wfp_zenodo}
\bibinfo{author}{Martini, G.} \emph{et~al.}
\newblock \bibinfo{journal}{\bibinfo{title}{Data and code for machine learning can guide food security efforts when primary data is not available}}.
\newblock {\emph{\JournalTitle{Zenodo}}} \url{https://doi.org/10.1101/2021.06.23.21259419} (\bibinfo{year}{2022}).

\bibitem{api_hungermap}
\bibinfo{author}{WFP}.
\newblock \bibinfo{title}{Hunger map wfp api}.
\newblock \bibinfo{howpublished}{Available at \url{https://static.hungermapdata.org/api-catalog/}}.

\bibitem{ipcmanual3}
\bibinfo{author}{IPC-Manual}.
\newblock \bibinfo{title}{Technical manual version 3.1}.
\newblock \bibinfo{howpublished}{\url{https://www.ipcinfo.org/fileadmin/user_upload/ipcinfo/manual/IPC_Technical_Manual_3_Final.pdf}} (\bibinfo{year}{2023}).
\newblock \bibinfo{note}{Accessed: 2023-10-18}.

\bibitem{hoddinott2002dietary}
\bibinfo{author}{Hoddinott, J.} \& \bibinfo{author}{Yohannes, Y.}
\newblock \emph{\bibinfo{title}{Dietary diversity as a household food security indicator}} (\bibinfo{publisher}{Food and Nutrition Technical Assistance Project (FANTA), Academy for~…}, \bibinfo{year}{2002}).

\bibitem{Leroy2015}
\bibinfo{author}{Leroy, J.~L.}, \bibinfo{author}{Ruel, M.}, \bibinfo{author}{Frongillo, E.~A.}, \bibinfo{author}{Harris, J.} \& \bibinfo{author}{Ballard, T.~J.}
\newblock \bibinfo{journal}{\bibinfo{title}{Measuring the food access dimension of food security: a critical review and mapping of indicators}}.
\newblock {\emph{\JournalTitle{Food and nutrition bulletin}}} \textbf{\bibinfo{volume}{36}}, \bibinfo{pages}{167--195} (\bibinfo{year}{2015}).

\bibitem{infoth}
\bibinfo{author}{Cover, T.} \& \bibinfo{author}{Thomas, J.}
\newblock \emph{\bibinfo{title}{Elements of Information Theory}} (\bibinfo{publisher}{Wiley}, \bibinfo{year}{2006}).

\end{thebibliography}

\section*{Acknowledgements} 
We thank the World Food Program, in particular Kusum Hachhethu, Duccio Piovani and Giulia Martini for data provision and consultancy on the manuscript preparation and revision. 

\section*{Author contributions statement}

M.M. and M.R. conceived the experiment(s), M.M. and M.R. conducted the experiment(s),
M.A., M.M. and M.R. conducted the visualisation(s),
M.M., M.R. and A-C.T. analysed the results. All authors reviewed the manuscript.

\section*{Figures \& Tables}

\end{document}